\documentclass[11pt,a4paper]{article}

\usepackage[utf8]{inputenc}
\usepackage[T1]{fontenc}
\usepackage[english]{babel}
\usepackage{lmodern}
\usepackage{microtype}
\usepackage{graphicx}
\usepackage{booktabs}
\usepackage{tabularx}
\usepackage{array}
\usepackage{amsmath}
\usepackage{url}
\usepackage{xurl}
\usepackage{xcolor}
\usepackage[hidelinks]{hyperref}
\usepackage{geometry}
\usepackage{caption}
\usepackage{enumitem}
\setlist{nosep,leftmargin=1.4em}
\usepackage{csquotes}
\usepackage{siunitx}
\usepackage{tikz}
\usetikzlibrary{positioning,fit,backgrounds}
\usepackage{placeins}

\newcommand{\EUR}[1]{EUR\,#1}

\newcolumntype{Y}{>{\raggedright\arraybackslash}X}

\title{\textbf{From 50K to 8.2 Million in 24 Hours:\\
Vozinha's Algorithmic Consecration and the Multilingual\\
Making of World Cup Visibility}}

\author{%
  Vinicius Covas\thanks{Research Professor, Faculty of Communication,
  Universidad An\'ahuac M\'exico; research areas include communication,
  artificial intelligence, platform visibility, and human/nonhuman
  communication.}\\
  Faculty of Communication, Universidad An\'ahuac M\'exico\\
  \texttt{vinicius.covas@anahuac.mx}\\
  ORCID: 0000-0001-9948-2940
}

\date{June 16, 2026}

\begin{document}
\maketitle

% ============================================================
\begin{abstract}
\noindent
We present a \emph{multilingual computational discourse analysis} of how language
constructed the algorithmic consecration of Vozinha, the 40-year-old Cape Verde
goalkeeper, after Spain~0--0 Cape Verde at the 2026 FIFA World Cup. Our central
contribution is computational-linguistic: a multilingual corpus (Portuguese,
Spanish, English, French), a nine-frame narrative taxonomy with cue-based
\emph{frame annotation}, a reproducible annotation pipeline combining
LLM-assisted suggestion with human validation, and an analysis of
\emph{cross-lingual narrative diffusion} across discourse phases. We treat the
platform follower count itself---narrated as \enquote{50k to 8M}---as a
\emph{linguistic object}: a circulating, narratable proof of visibility rather
than a mere measurement. We analyze how recurring frames---underdog heroism,
national history, age and late recognition, platform-metric spectacle, Brazilian
live mobilization, Spanish crisis, and affective appeals---diffused across
languages and time. The follower-growth timeline is used only as \emph{contextual
metadata}: we reconstruct a conservative phase structure, not a continuous
API-native series, and we type every datapoint by value class, confidence, and
evidence type. The only exact primary scraper anchor is \num{8235652} followers at
2026-06-16~15:47~UTC; all other figures are reported as estimated ranges or
thresholds (e.g., an estimated pre-match baseline of 45k--56k). Our findings
suggest that distinct languages carried distinct frames---Portuguese mobilization,
Spanish crisis, English nation-making---and that platform-metric language was the
shared mechanism converting a peripheral athletic performance into global symbolic
visibility. We release the corpus schema, frame taxonomy, annotation guidelines, a
hashed visual-evidence log, and a typed timeline. As a v0.1 pilot, we do not claim
a complete follower history; full double-annotation and inter-annotator agreement
are flagged as planned work.


\medskip
\noindent\textbf{Version note.} This is a v0.1 pilot preprint. It introduces the
case, corpus schema, frame taxonomy, annotation procedure, and a reconstructed
timeline as contextual metadata. A later version will expand the corpus, complete
double annotation, and report inter-annotator agreement.
\end{abstract}

% ============================================================
\section{Introduction}
On 15 June 2026, in Atlanta, Cape Verde held Spain to a 0--0 draw at the FIFA
World Cup. The goalkeeper, Josimar José Évora Dias---\enquote{Vozinha}
(\texttt{@vozinha1})---produced a high-rated defensive performance, and within
roughly a day his Instagram following rose from a reported pre-match baseline in
the tens of thousands to several million. A user-run Apify collection records
exactly \num{8235652} followers at 2026-06-16~15:47~UTC, which we adopt as our
single primary anchor.

The empirical puzzle is not that an athlete gained followers, but \emph{how}
language across four linguistic publics transformed a peripheral 0--0 draw into a
globally legible event of symbolic consecration. We argue that Vozinha was not
merely \emph{viral} but \emph{linguistically and algorithmically consecrated}:
narrative frames---and, crucially, language \emph{about the metric itself}---did
the work of converting an athletic act into durable visibility.

Several features make this case unusually legible as a discourse object. Cape
Verde is a small, peripheral footballing nation whose presence at a World Cup was
itself narratable as history; its goalkeeper was a 40-year-old who had reportedly
turned professional only at 25 and had played much of his career in lower divisions
and outside Europe. The match was carried to a very large Lusophone audience by an
internet-native broadcast (CazéTV, hosted by the streamer Casimiro), so that the
performance and the appeal to \enquote{follow him} reached millions synchronously.
The result was a dense, multilingual stream of public discourse---Portuguese,
Spanish, English, and French---produced within hours, in which the same event was
encoded very differently depending on the language and the audience.

Within that stream, the follower metric did not stay in the background. It moved to
the foreground and became the headline: \enquote{50k to 8M} was repeated as a
self-evident proof that something historic had happened. Before/after screenshots
of the profile, cross-sport comparisons (\enquote{more followers than Wembanyama}),
and rolling counts circulated as public evidence. In other words, the number became
a unit of \emph{public proof}---a way of saying \enquote{this matters} that is
legible across languages without translation. This is the phenomenon we call, in
shorthand, the multilingual making of World Cup visibility: a peripheral athletic
performance is converted into platformed symbolic recognition through the joint work
of language and metric.

\paragraph{Contribution (cs.CL).}
This is a computational-linguistics study, not a sports-marketing one. The object
of study is language. We contribute: (i) a multilingual corpus of news leads,
social posts, and captions in Portuguese, Spanish, English, and French; (ii) a
nine-frame narrative taxonomy (F1--F9) with cue-based assignment rules; (iii) a
reproducible LLM-assisted-plus-human annotation pipeline; and (iv) a cross-lingual
analysis of frame distribution and temporal diffusion. The follower timeline is
used only to anchor discourse phases. We operationalize \emph{algorithmic
consecration} through observable linguistic frames rather than asserting it, and
we deliberately type and bound every quantitative claim.

% ============================================================
\section{Related Work}
Our work sits at the intersection of computational discourse analysis,
multilingual framing, and narrative diffusion. Frame analysis treats discourse as
selecting and foregrounding aspects of a perceived reality; computational framing
research operationalizes frames through lexical and rhetorical cues amenable to
annotation. Cross-lingual NLP has shown that the \emph{same} event can be framed
divergently across languages, motivating our comparison of Portuguese, Spanish,
English, and French coverage of one event.

Theoretically, we draw on the attention economy
\cite{goldhaber1997attention,davenport2001attention}, the culture of connectivity
and algorithmic power \cite{vandijck2013culture,bucher2018ifthen}, microcelebrity
and internet fame \cite{marwick2013status,abidin2018internet}, and the reputation
economy \cite{hearn2010structuring,gandini2016reputation}. The platformization of
sport \cite{fujak2026influencer} contextualizes how athletic performance is
recoded as influencer value. From the computational-social-science side, work on
virality and coordinated amplification
\cite{virality2023measuring,pacheco2020coordinated,coordinated2023campaigns}
informs how we treat volatile engagement signals---as discursive evidence, not as
metric ground truth. Our distinctive move is \emph{metric-as-discourse}: we treat
the follower count as a linguistic object whose narration (\enquote{50k to 8M})
is itself a frame.

% ============================================================
\section{Case Context}
\paragraph{Cape Verde, Vozinha, and a peripheral debut.}
Cape Verde---an archipelago nation of roughly half a million people---reached the
2026 FIFA World Cup for the first time, a fact that international and Lusophone
coverage repeatedly narrated as historic. Its goalkeeper, Josimar José Évora Dias
(\enquote{Vozinha}), is a 40-year-old whose listed transfer-market value was modest
(reported around \EUR{50{,}000}) \cite{transfermarkt_vozinha,as_vozinha_value} and
who, by widely circulated biographical accounts, turned professional comparatively
late and built his career largely outside the European elite, including spells in
lower divisions. This biography is
itself a discourse resource: it supplies the underdog, veteran, and
late-recognition frames that recur across languages.

\paragraph{The match and the live mobilization.}
Against Spain---a tournament favourite---Cape Verde earned a 0--0 draw, with
Vozinha producing a high-rated defensive performance (multiple saves) that
match-data providers scored well \cite{sofascore_vozinha}. Crucially, the match reached an enormous
Lusophone audience through CazéTV, the internet-native broadcast hosted by the
Brazilian streamer Casimiro. Portuguese-language coverage frames a live, explicit
appeal to the audience---a \emph{mutirão} (collective drive) to follow
\texttt{@vozinha1}---as the proximate engine of the initial surge, with the count
reportedly moving into the hundreds of thousands within minutes of the appeal. This
live-broadcast mobilization is what distinguishes the case from a pre-event
influencer campaign and is central to why the surge was both fast and narratable.

\paragraph{Why \enquote{50k to 8.2M} became the hook.}
Because the pre-match baseline was small and widely reported (an estimated 45k--56k,
with one cold baseline of about 31.5k roughly a week earlier) and the post-match
ceiling was large and primary-anchored (\num{8235652}), the gap was easy to state
and easy to share. The number, not the save, became the lead. This is the empirical
hook of the paper and the reason the metric must be analyzed as a linguistic object
rather than only as a measurement.

\paragraph{Harmonized time axis and data discipline.}
We harmonize all times on a single axis: \textbf{kickoff 2026-06-15~16:00~UTC},
\textbf{full-time 2026-06-15~17:52~UTC}. We do \emph{not} claim a continuous
follower time series. Instead, the conservative timeline (Table~\ref{tab:timeline})
indicates the following phases: an estimated pre-match baseline of roughly 45k--56k;
a during-match move into the hundreds of thousands after the CazéTV mobilization; a
crossing of 1M in the full-time/dressing-room window; \enquote{over 2M} the same
day \cite{reuters_vozinha}; 6M+ the next morning \cite{guardian_vozinha}; and the
primary Apify anchor of \num{8235652} at 2026-06-16~15:47~UTC. The live mobilization
is documented in Brazilian coverage \cite{terra_cazetv_vozinha,uol_vozinha}, and the
threshold figures are corroborated across international outlets
\cite{straitstimes_vozinha,qazinform_vozinha,leadership_vozinha,yahoo_vozinha8m,espn_vozinha}.
We treat reported journalistic figures as thresholds or
ranges; the only exact value is the primary scraper anchor. Source-side timezone
slips (e.g., a \enquote{3:45pm EST} figure during a period that is EDT) are
preserved as notes rather than silently corrected. We also note an instrumentation
caveat: a third-party analytics service displayed a stale cached value during the
peak, which we record as an observation about measurement under load rather than as
a curve point.

\begin{table}[htbp]
  \centering\footnotesize
  \caption{Conservative paper-ready timeline used as contextual metadata. Values are
  typed as ranges, thresholds, reported values, or exact scraper values. Only the
  Apify anchor is exact. Harmonized axis: kickoff 16:00~UTC, full-time 17:52~UTC.}
  \label{tab:timeline}
  \begin{tabularx}{\linewidth}{@{}l Y l l l@{}}
    \toprule
    \textbf{Phase} & \textbf{Followers} & \textbf{Value type} & \textbf{Conf.} & \textbf{Evidence} \\
    \midrule
    pre-match baseline      & 45k--56k (range)        & range/estimate & med.-high & news-consensus \\
    during-match (CazéTV)   & 200k--300k (threshold)  & threshold      & medium    & news-reported \\
    full-time / dressing rm.& $\sim$1.0--1.1M         & threshold      & med.-high & news-reported \\
    same day (15 Jun)       & over 2M                 & threshold      & med.-high & news-reported \\
    night 15 / early 16 Jun & $\sim$5.0--5.7M         & reported       & medium    & news-reported \\
    morning 16 Jun          & 6.0--6.4M               & reported       & med.-high & news-reported \\
    early afternoon 16 Jun  & 7.6M                    & reported       & med.-high & news-reported \\
    primary anchor (Apify)  & \num{8235652}           & \textbf{exact} & very high & primary-scraper \\
    \bottomrule
  \end{tabularx}
\end{table}

% ============================================================
\section{Corpus and Annotation}
\paragraph{Sources and languages.}
The corpus comprises document-level excerpts---headlines, lead paragraphs, tweets,
Instagram story captions, and screenshot text---drawn from public news outlets and
public posts in Portuguese (\texttt{pt}), Spanish (\texttt{es}), English
(\texttt{en}), and French (\texttt{fr}). Each excerpt retains the original-language
\texttt{text\_excerpt} and an English working gloss (\texttt{translation\_en});
glosses are explicitly not certified translations, and claims rest on
original-language cues. The released schema
(\texttt{multilingual\_corpus\_schema.csv}) records language, timestamp, frame
labels, affective register, national reference, metric reference, platform
reference, and a confidence level per excerpt.

\paragraph{Confidence and evidence typing.}
Every excerpt carries a \texttt{confidence\_level} (very high $\rightarrow$ low),
and we distinguish evidence types: news consensus, news-reported, primary-scraper,
and screenshot. Screenshots are SHA-256 hashed for integrity and are never upgraded
to API-equivalent status. Volatile X engagement (likes, reposts, views) is used
only as discursive example, not as metric evidence.

\paragraph{Timeline as metadata.}
We separate a defensible \texttt{paper\_ready\_timeline} (points eligible for the
main figure) from a \texttt{supporting\_evidence\_timeline} (granular but weaker
points). Only the typed, conservative timeline anchors the discourse phases.

\paragraph{Visual evidence corpus.}
In addition to text excerpts, we logged a corpus of 48 screenshots captured on
2026-06-16 between approximately 09:17 and 09:28 local machine time: 9 from
X/Twitter and 39 from Instagram Stories. These images are valuable because they
show, in situ, the author, an absolute or relative timestamp, the displayed metric,
and---in several cases---the \texttt{@vozinha1} profile header itself. Three
independent profile snapshots are especially informative for the curve: a side-by-side
\enquote{50k$\rightarrow$1.9M} before/after panel (re-shared by the football account
@433), a \enquote{432 posts $\cdot$ 1.9M} header captured around a FIFA press-conference
moment, and an \enquote{8M} profile header captured the next morning. The X subset is
dominated by cross-sport comparison posts (placing Vozinha's count alongside athletes
such as Wembanyama, Mahomes, Stafford, Brady, the Yankees, and the Miami Heat) and by
official/large-account amplification (the official @fifaworldcup account, @433, and
the Spanish outlet @marca); the phenomenon was itself narrated as a platform event
in Brazilian coverage \cite{veja_cazetv}. Screenshots are treated as \emph{visual-discursive
evidence} and integrity-checked through SHA-256 hashes (released in
\texttt{evidence\_hashes.csv}). They are \emph{not} treated as API-equivalent
measurements: they confirm that values and discourse were publicly displayed, but
they do not prove that platform-side metrics were unedited and they do not yield a
dense time series. Relative Story timestamps (\enquote{18h ago}) are used for ordering
only, with a $\pm$1h margin, and the X timestamps are recorded as shown in the
user-interface timezone, to be normalized rather than assumed to be UTC.

% ============================================================
\section{Methods}
\paragraph{Frame taxonomy.}
We define nine narrative frames (Table~\ref{tab:frames}). A frame is assigned only
when a lexical or rhetorical \emph{cue} is present; assignment by inferred author
intent is disallowed. Annotators may assign multiple frames per excerpt.

\begin{table}[htbp]
  \centering\small
  \caption{Frame taxonomy (F1--F9). A frame is assigned only on an explicit
  lexical/rhetorical cue.}
  \label{tab:frames}
  \begin{tabularx}{\linewidth}{@{}l Y l@{}}
    \toprule
    \textbf{ID} & \textbf{Frame} & \textbf{Typical languages} \\
    \midrule
    F1 & underdog / peripheral hero & EN, ES, FR, PT \\
    F2 & national visibility / Cape Verde making history & EN, PT, ES \\
    F3 & age / veteran / late recognition & EN, ES, PT \\
    F4 & platform metric spectacle (\enquote{50k to 8M}) & EN, PT, ES, FR \\
    F5 & Brazilian live mobilization / CazéTV & PT \\
    F6 & Spanish crisis / frustration & ES \\
    F7 & affect / family / tears & ES, PT, EN \\
    F8 & economic attention / influencer value & EN, ES, PT \\
    F9 & cross-sport comparison & EN, PT \\
    \bottomrule
  \end{tabularx}
\end{table}

\paragraph{Annotation pipeline (LLM-assisted, human-validated).}
A language model proposes frame labels and a gloss for each excerpt; this is a
\emph{suggestion}, never the final label. A human annotator confirms or overrides
each label against the cue rules, logging overrides. Disagreements are adjudicated;
unresolved items are marked low-confidence and excluded from main claims. We
report planned inter-annotator agreement (Cohen's $\kappa$ / Krippendorff's
$\alpha$) on a double-annotated subset; the current release is a seeded sample for
schema demonstration. The present v0.1 release is a seeded pilot corpus intended to
validate the taxonomy, annotation schema, and temporal framing procedure. It should
not be interpreted as a representative sample of all coverage or social media
discourse around the event.

\begin{figure}[htbp]
  \centering
  \includegraphics[width=0.92\linewidth]{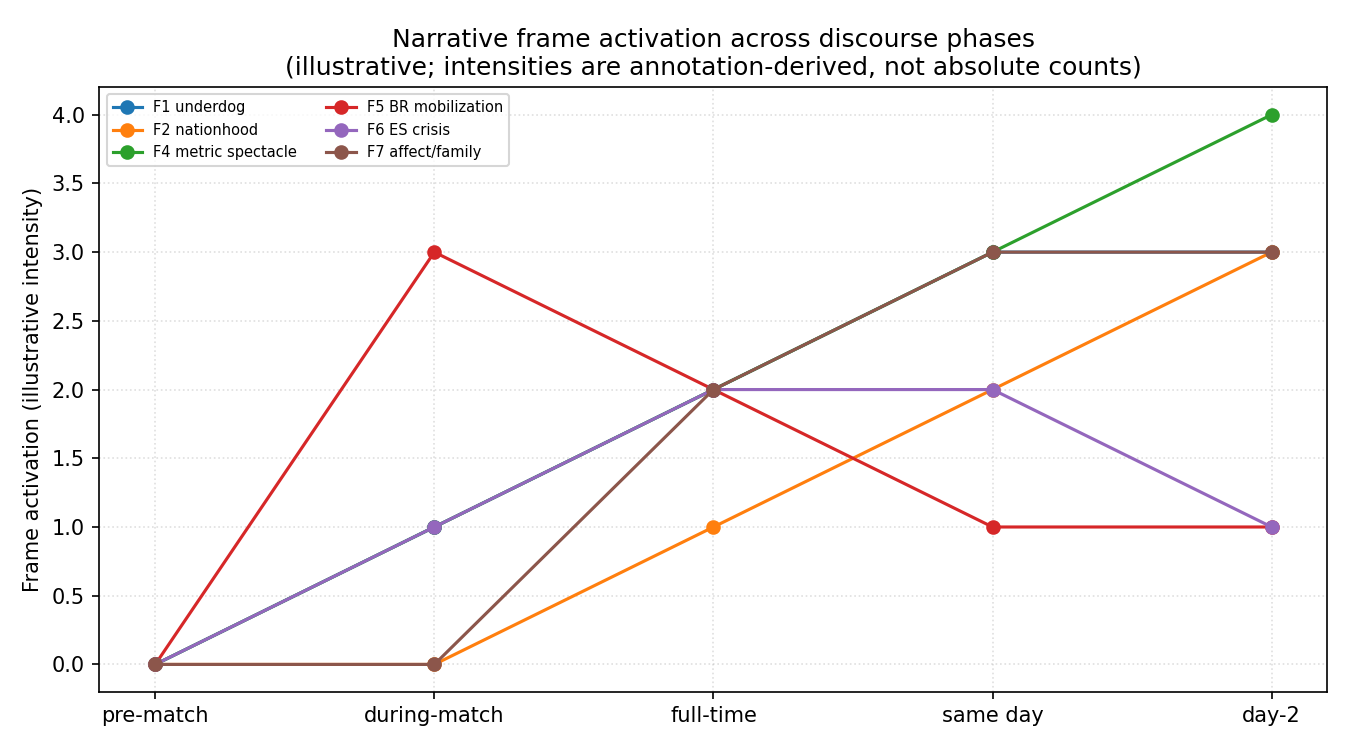}
  \caption{Pilot-coded frame activation: temporal diffusion of narrative frames
  (F1--F9) across the harmonized discourse phases (during-match $\rightarrow$
  day-2). This is a descriptive visualization of the seeded v0.1 corpus and should
  not be interpreted as a population-level estimate.}
  \label{fig:diffusion}
\end{figure}

\begin{figure}[htbp]
  \centering
  \includegraphics[width=0.78\linewidth]{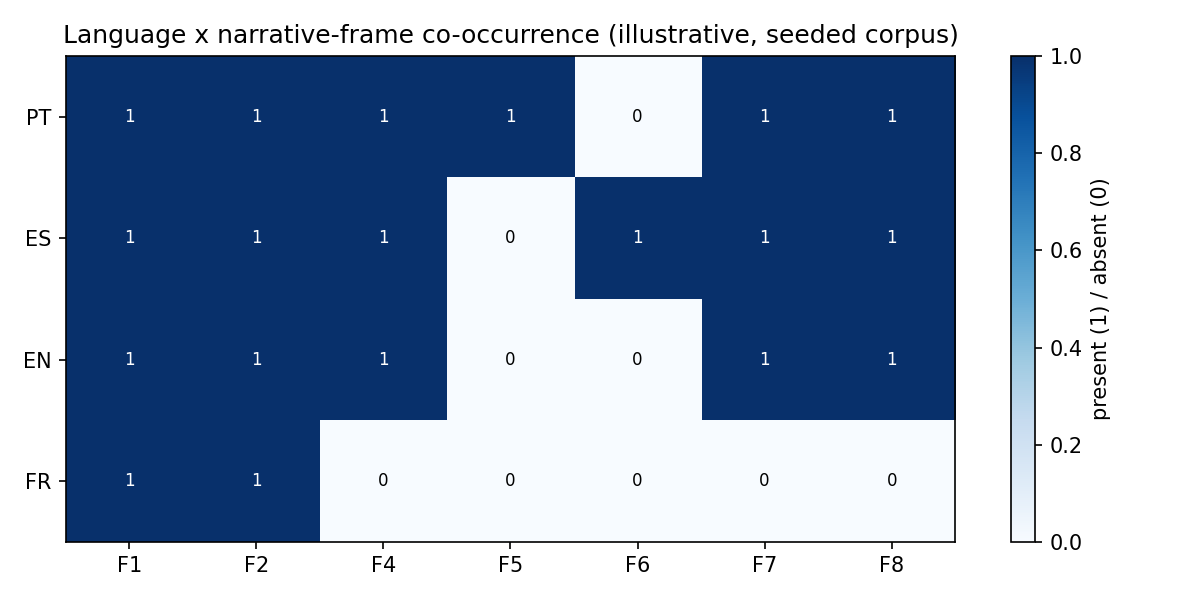}
  \caption{Language $\times$ frame presence in the seeded v0.1 corpus, illustrating
  that different languages carry different frames (Portuguese mobilization,
  Spanish crisis, English nation-making, with the metric-spectacle frame common
  across languages). Cells indicate binary presence/absence in the seeded corpus,
  not frequency or intensity, and should not be read as population-level estimates.}
  \label{fig:heatmap}
\end{figure}

\paragraph{Extraction and temporal diffusion.}
We apply keyword/phrase extraction and named-entity extraction (e.g., \emph{Cape
Verde}, \emph{Spain}, \emph{Brazil}, \emph{CazéTV}, \emph{Casimiro}) and track the
activation of each frame across the harmonized discourse phases to characterize
cross-lingual diffusion (Figures~\ref{fig:diffusion} and~\ref{fig:heatmap}).

\paragraph{Screenshot logging and visual-text extraction.}
For the visual evidence corpus we follow a deliberately conservative procedure.
Each screenshot is stored as a file and hashed with SHA-256; the hash and filename
are recorded in \texttt{evidence\_hashes.csv} so that any later copy can be verified
as identical to the logged artifact. The visible text in each image---author handle,
displayed timestamp, follower value, and any caption---was read and transcribed
manually into a structured log (\texttt{visual\_evidence\_log.md}); we did not rely
on automated OCR, in order to avoid silent transcription errors on stylized
platform UI. From that transcription, the same cue-based rules used for the text
corpus were applied to assign frame labels (for example, a cross-sport comparison
image is coded F9; a profile-count before/after panel is coded F4). Engagement
numbers visible in the images (likes, reposts, views) were recorded for context but
are explicitly \emph{not} used as metric evidence, because they are volatile and
were not collected through a logged API call.

\paragraph{Why screenshots differ from scraper/API data.}
We keep a strict separation between two evidence types. A scraper/API reading (such
as the primary Apify anchor) is a value obtained through a logged, timestamped
collection that can in principle be re-run and audited. A screenshot is a record of
what a platform interface \emph{displayed} to a user at capture time. Screenshots are
strong evidence that a value and a piece of discourse were public, and they are
well-suited to discourse analysis; but they cannot establish a continuous series and
cannot rule out interface- or client-side artifacts. We therefore treat the
screenshots as linguistic/visual evidence about how the metric circulated, and the
Apify anchor as the only exact quantitative point.

% ============================================================
\section{Results}
Representative cross-lingual cues are summarized in Table~\ref{tab:examples}.

\begin{figure}[htbp]
  \centering
  \includegraphics[width=0.92\linewidth]{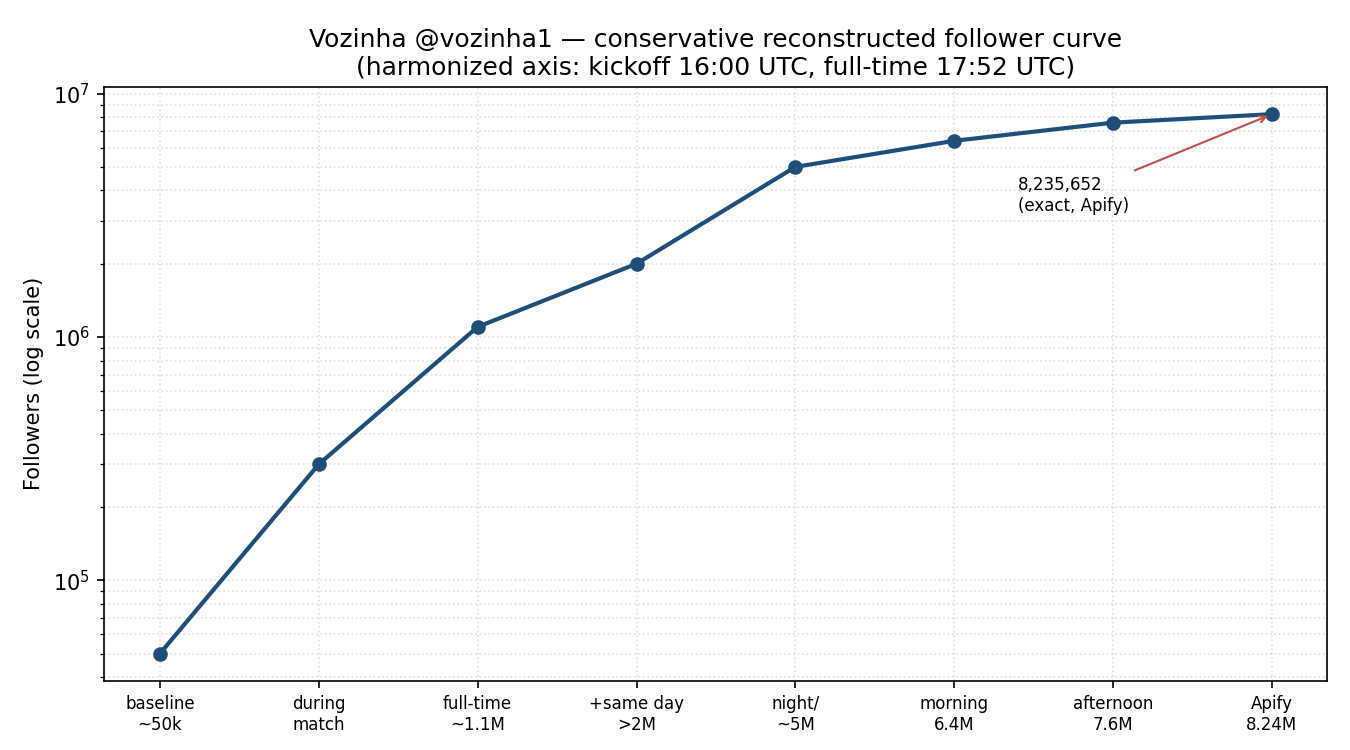}
  \caption{Conservative follower-growth timeline (contextual metadata), log scale.
  Points are typed by value class and confidence; only \texttt{paper\_ready}
  points eligible for the main figure are shown. The single exact point is the
  primary Apify anchor (\num{8235652} at 2026-06-16~15:47~UTC). We do not claim a
  continuous series.}
  \label{fig:timeline}
\end{figure}

\begin{table}[htbp]
  \centering\footnotesize
  \caption{Representative frame examples across languages. Excerpts are short
  cue phrases (paraphrased or abbreviated to avoid reproducing long copyrighted
  passages); the full corpus schema records original-language text and glosses.}
  \label{tab:examples}
  \begin{tabularx}{\linewidth}{@{}l l Y l l@{}}
    \toprule
    \textbf{Lang} & \textbf{Source type} & \textbf{Cue / short excerpt} & \textbf{Frame} & \textbf{Conf.} \\
    \midrule
    pt & broadcast / social & \emph{mutir\~ao} to follow; \enquote{os brasileiros} & F5; F4 & high \\
    es & news lead         & \enquote{le hace la vida imposible a Espa\~na}        & F6; F8 & high \\
    en & news headline     & \enquote{unbeatable goalkeeper}; \enquote{making history} & F1; F2 & med.-high \\
    en & news (affect)     & mother could not afford visa fees                    & F7; F1 & med.-high \\
    fr & news lead         & feat $\rightarrow$ network visibility                & F1; F2 & low-med. \\
    multi & metric spectacle & \enquote{50k to 8M}; before/after profile          & F4     & high \\
    en & X comparison      & \enquote{more followers than Wembanyama / Brady}     & F9; F4 & medium \\
    \bottomrule
  \end{tabularx}
  \par\smallskip
  {\scriptsize Interpretation: Portuguese carries live mobilization, Spanish carries
  crisis/economic framing, English carries underdog/nation-making and affect, French
  marks the passage from feat to network visibility, and the metric-spectacle frame
  is shared across languages. Evidence type is news/social text except the comparison
  row (screenshot, visual-discursive).}
\end{table}

\paragraph{R1: Metric-spectacle language (\enquote{50k to 8M}) as public proof.}
Across languages, the follower count is repeatedly foregrounded as the headline
itself (frame F4) \cite{yahoo_vozinha8m,qazinform_vozinha,mybettingsites_gainers,instastatistics_vozinha}.
The metric is narrated as evidence of significance---a self-referential proof in
which the number, rather than the save, becomes the news. Before/after profile
screenshots (e.g., 50k$\rightarrow$1.9M) circulate as visual instantiations of this
frame.

\paragraph{R2: Portuguese/Brazilian discourse as mobilization language.}
Portuguese-language coverage is distinguished by mobilization vocabulary tied to
CazéTV/Casimiro (frame F5) \cite{terra_cazetv_vozinha,uol_vozinha}: explicit calls
to follow, the framing of a collective \emph{mutirão}, and gratitude posts addressed
to \enquote{os brasileiros.} Here language does not merely describe the surge; it
enacts it.

\paragraph{R3: Spanish discourse as crisis/frustration frame.}
Spanish-language coverage foregrounds Spain's failure to score (frame F6)
\cite{elpais_vozinha,as_vozinha_value}: \enquote{le hace la vida imposible a España},
framing the 0--0 as a Spanish embarrassment more than a Cape Verde triumph,
frequently co-occurring with the economic-value frame (F8, the \EUR{50{,}000}
keeper).

\paragraph{R4: English/global discourse as underdog/nation-making frame.}
English-language coverage privileges the underdog and nationhood frames (F1, F2,
F3) \cite{straitstimes_vozinha,espn_vozinha}: the \enquote{unbeatable goalkeeper},
the veteran \enquote{at 40}, and Cape Verde \enquote{making history.} Affective
family framing (F7)---the mother who could not afford visa fees
\cite{indianexpress_vozinha}---amplifies shareability.

\paragraph{R5: Platform language converts performance into consecration.}
Synthesizing R1--R4, the cross-lingual evidence suggests that platform-metric
language is the connective tissue: each linguistic public supplies a distinct
frame, but all converge on the metric as the durable, portable proof. The
performance becomes consecrated not when it ends but when it is \emph{counted} and
narrated as a count.

\subsection{Metric screenshots as linguistic objects}
The visual evidence corpus makes the metric-as-discourse argument concrete. Across
the screenshots, the follower count is not used as a quiet measurement; it is put to
work as a sign. It appears as before/after proof (the side-by-side
\enquote{50k$\rightarrow$1.9M} profile panel), as headline (the \enquote{8M} profile
header reproduced as the point of a post), as meme and comparison device (the
cross-sport posts that line Vozinha's count up against Wembanyama, Stafford, Brady,
the Yankees, and the Miami Heat), and as a legitimacy signal (official and large
accounts re-circulating the count as confirmation that the moment is historic). In
each register, the number is doing rhetorical work that ordinary language would
otherwise have to do: it asserts significance in a form that travels across
languages without translation. A count is legible to Portuguese, Spanish, English,
and French publics simultaneously, which is precisely why it becomes the shared
object on which otherwise divergent frames converge.

This is the central interpretive claim of the paper: \emph{the metric did not merely
record visibility; it became one of the primary languages through which visibility
was produced.} Treated this way, the follower count is not the outcome variable of a
marketing process but a linguistic object with its own pragmatics---a way of saying
\enquote{this matters} that is quotable, screenshot-able, and comparable, and that
therefore circulates as discourse in its own right.

\begin{figure}[htbp]
  \centering
  \footnotesize
  \begin{tikzpicture}[
    node distance=4mm and 4mm,
    panel/.style={draw, rounded corners=2pt, align=left, inner sep=4pt,
      text width=0.28\linewidth, font=\scriptsize, minimum height=15mm},
  ]
    \node[panel] (a) {\textbf{(A) Profile before/after}\\
      Source: @433 (IG Story)\\
      Snapshot: 50k $\rightarrow$ 1.9M\\
      Frame: F4 metric spectacle\\
      Cue: side-by-side count};
    \node[panel, right=of a] (b) {\textbf{(B) Peak profile header}\\
      Source: \texttt{@vozinha1} header\\
      Snapshot: 439 posts $\cdot$ 8M\\
      Frame: F4 metric spectacle\\
      Cue: count as headline};
    \node[panel, right=of b] (c) {\textbf{(C) Cross-sport comparison}\\
      Source: X comparison post\\
      \enquote{$>$ Brady / Wembanyama}\\
      Frame: F9 cross-sport\\
      Cue: count vs.\ NBA/NFL};
    \node[panel, below=of a] (d) {\textbf{(D) Official amplification}\\
      Source: @fifaworldcup / @433\\
      \enquote{making history}\\
      Frame: F2 nation-making\\
      Cue: institutional repost};
    \node[panel, below=of b] (e) {\textbf{(E) Live mobilization}\\
      Source: CazéTV (PT)\\
      \emph{mutir\~ao} to follow\\
      Frame: F5 mobilization\\
      Cue: explicit call to follow};
    \node[panel, below=of c] (f) {\textbf{(F) Affect / national pride}\\
      Source: peers / fans (PT)\\
      tears, \enquote{legend}\\
      Frame: F7 affect / family\\
      Cue: emotional register};
  \end{tikzpicture}
  \caption{Selected visual-discursive evidence from the v0.1 corpus, presented as a
  schematic montage (source type, displayed value, frame, and visual cue) rather
  than as reproduced screenshots. Screenshots are used as public visual evidence of
  discourse and metric circulation; they are \emph{not} treated as API-equivalent
  measurements. Full evidence descriptions and SHA-256 hashes are archived in the
  data package.}
  \label{fig:montage}
\end{figure}

% ============================================================
\section{Discussion}
We define \textbf{algorithmic consecration} as a process that is simultaneously
\emph{language-mediated} and \emph{platform-mediated}: symbolic value is conferred
when public discourse encodes an event in shareable frames \emph{and} when a
platform metric renders that value legible and countable. \enquote{Viral fame} is
insufficient as a description because it foregrounds spread while ignoring the
linguistic labor that makes spread meaningful and the metric labor that makes it
durable.

The case also illustrates \emph{peripheral visibility} and nationhood: a small
nation and a low-market-value athlete acquire global symbolic presence through a
cross-lingual frame assemblage. The Portuguese mobilization frame seeds the surge;
the Spanish crisis frame supplies conflict; the English nation-making frame supplies
narrative durability; and the metric-spectacle frame, common to all, converts the
moment into a portable number.

% ============================================================
\section{Limitations and Ethics}
\paragraph{Limitations.}
We reconstruct a conservative phase timeline; we do \emph{not} possess a complete
API-native follower history, and we treat journalistic figures as thresholds or
ranges rather than exact values. The only exact point is the primary Apify anchor.
The corpus is a seeded v0.1 sample for schema demonstration; full double-annotation
and inter-annotator agreement are planned work. The present v0.1 release is a seeded
pilot corpus intended to validate the taxonomy, annotation schema, and temporal
framing procedure. It should not be interpreted as a representative sample of all
coverage or social media discourse around the event. As a single case study,
generalizability is limited; we position the contribution as a method and typology,
with comparable cases (e.g., pre-event influencer-driven surges at the same
tournament \cite{yahoo_payne}) available as anchors.

Several limitations are specific to the visual evidence corpus. First, copyright and
platform constraints limit the reproduction of raw screenshots; rather than embed
third-party images, we describe them, hash them, and present a schematic montage
(Figure~\ref{fig:montage}), with full descriptions archived in the data package.
Second, screenshots are artifacts of what an interface displayed, not complete
records: they cannot establish a dense series and cannot exclude client- or
interface-side rendering effects. Third, the capture times reflect the local
machine and the platform user-interface timezone (and, for Stories, rounded relative
timestamps such as \enquote{18h ago}); these are used for ordering, not as absolute
UTC, and are flagged for normalization. Fourth, the visual corpus is selectively
captured rather than systematically sampled, so frame frequencies in it are
illustrative, not representative. Future work should collect API-native post and
profile data, where legally and technically permitted, alongside the discourse
corpus, so that the linguistic analysis can be paired with an auditable metric
series.

\paragraph{Ethics.}
We use only public outlets, public posts, and public figures' accounts; no private
data is analyzed. Any X data would be released dehydrated (IDs) consistent with
platform terms. Translations are working glosses and marked as such.

% ============================================================
\section{Conclusion}
We have offered a multilingual computational discourse analysis of how language
consecrated Vozinha after Spain~0--0 Cape Verde. The evidence suggests that
distinct languages carried distinct frames and that platform-metric language was
the shared mechanism converting a peripheral performance into global symbolic
visibility. We release the corpus schema, frame taxonomy, annotation guidelines,
and a typed, conservative timeline to support replication and extension to further
cases of language-mediated algorithmic consecration.

% ============================================================
\FloatBarrier

% ============================================================
\section*{Data Availability}
The v0.1 dataset, source log, frame taxonomy, corpus schema, evidence log, and
reconstructed timeline are archived on Zenodo:
\url{https://doi.org/10.5281/zenodo.20722235}. A Kaggle exploratory mirror is
available at:
\url{https://www.kaggle.com/datasets/viniciuscovas/vozinha-instagram-surge}.
Zenodo remains the official citation target for the dataset and evidence package.

% ============================================================
% REFERENCES
% ============================================================
\bibliographystyle{unsrt}
\bibliography{references}

\end{document}